\documentclass{llncs}
\usepackage{makeidx}  % allows for indexgeneration
\usepackage{amsmath,amssymb}
\usepackage{algorithm}
\usepackage[noend]{algpseudocode}

\usepackage{graphicx}

\usepackage{subfigure}

\graphicspath{{./}{figures/}}

\makeatletter
\def\BState{\State\hskip-\ALG@thistlm}
\makeatother

\begin{document}

\title{Multiple-View Spectral Clustering for\\Group-wise Functional Community Detection}
\titlerunning{Multiple-view Spectral Clustering}  % abbreviated title (for running head)
\author{Nathan D. Cahill\inst{1,2}\thanks{Send correspondence to: \email{nathan.cahill@rit.edu}} \and Harmeet Singh\inst{1} \and Chao Zhang\inst{3,4} \and Daryl A. Corcoran\inst{1,2} \and Alison M. Prengaman\inst{1,2} \and Paul S. Wenger\inst{2} \and John F. Hamilton\inst{2} \and \\ Peter Bajorski\inst{2} \and Andrew M. Michael\inst{4}}
\authorrunning{N.~D. Cahill et al.} % abbreviated author list (for running head)
\institute{Image Computing \& Analysis Laboratory (ICAL), RIT, Rochester, NY, USA
\and
School of Mathematical Sciences, RIT, Rochester, NY, USA
\and
Center for Imaging Science, RIT, Rochester, NY, USA
\and
Autism and Developmental Medicine Institute,\\Geisinger Health System, Danville, PA, USA
}

\maketitle              % typeset the title of the contribution
\thispagestyle{plain}

\vspace{60pt}

\begin{abstract} 
Functional connectivity analysis yields powerful insights into our understanding of the human brain. Group-wise functional community detection aims to partition the brain into clusters, or communities, in which functional activity is inter-regionally correlated in a common manner across a group of subjects. 
In this article, we show how to use multiple-view spectral clustering to perform group-wise functional community detection. In a series of experiments on 291 subjects from the Human Connectome Project, we compare three versions of multiple-view spectral clustering: MVSC (uniform weights), MVSCW (weights based on subject-specific embedding quality), and AASC (weights optimized along with the embedding) with the competing technique of Joint Diagonalization of Laplacians (JDL). Results show that multiple-view spectral clustering not only yields group-wise functional communities that are more consistent than JDL when using randomly selected subsets of individual brains, but it is several orders of magnitude faster than JDL.
\end{abstract}

%%%%%%%%%%%%%%%%%%%%%%%%%%%%%%%%%%%%%%%%%%%%%%%%%%%%%%%%%%%%%%%%
%%%%%%%%%%%%%%%%%%%%%%%%%%%%%%%%%%%%%%%%%%%%%%%%%%%%%%%%%%%%%%%%
\section{Introduction}
\label{sec:intro}

In recent years, a variety of ideas have been proposed around the use of spectral clustering techniques to fuse group-wise information from structural and/or functional brain imagery. Van den Heuvel et al. \cite{vandenheuvel2008ncg} use a two-stage approach for identifying functional networks from resting-state fMRI (rsfMRI) that first clusters individual subjects and then partitions a graph constructed by connecting individual clusters across the group of subjects. Chen et al. \cite{chen2013igw} fuse diffusion tensor imagery (DTI) and rsfMRI of multiple subjects using a co-training algorithm \cite{kumar2011cta} that projects affinity matrices of individual subjects onto the eigenspaces of those of other subjects. While these approaches do yield useful results and insights, neither is stated as a well-posed optimization problem over the space of subject graphs, making them difficult to analyze from a mathematical perspective.

One recent approach that does formulate group-wise functional community detection as a well-posed optimization problem is described in Dodero et al. \cite{dodero2014gwf}: Joint Diagonalization of Laplacians (JDL) finds an optimal group-wise embedding of the graphs representing resting-state networks of each subject, and it subsequently applies a standard clustering technique to the optimal embedding in order to generate the functional communities. While JDL is a mathematically intriguing approach in its own right, the underlying algorithm \cite{cardoso1996jas} on which it relies requires computing an embedding in $\mathbb{R}^{n}$, where $n$ is the number of graph vertices. Since the desired number of clusters, $k$, is typically far less than $n$, JDL may be very inefficient compared to algorithms that would only require embeddings in $\mathbb{R}^{k-1}$. Furthermore, the objective function minimized by JDL is not related to any specific graph partitioning cost, making it difficult to infer the true relative quality between two different clustering results. 

In this article, we show how to use ideas from multiple-view spectral clustering \cite{huang2012aas,zhou2007sct} to design group-wise functional community detection algorithms that approximately minimize a well-defined cut cost and only require computing embeddings in $\mathbb{R}^{k-1}$. We then perform a series of experiments on a group of subjects from the Human Connectome Project \cite{hcpOnline} to show that multiple-view spectral clustering often yields more consistent clusters and is significantly more computationally efficient than JDL. 

%%%%%%%%%%%%%%%%%%%%%%%%%%%%%%%%%%%%%%%%%%%%%%%%%%%%%%%%%%%%%%%%
%%%%%%%%%%%%%%%%%%%%%%%%%%%%%%%%%%%%%%%%%%%%%%%%%%%%%%%%%%%%%%%%
\section{Preliminaries}
\label{sec:prelims}

Consider an undirected weighted graph $\mathcal{G} = \left(V,\mathcal{E}\right)$ having $n$ vertices that we wish to partition into $k$ disjoint subgraphs $\mathcal{G}_{i} = \left(V_{i},\mathcal{E}_{i}\right)$, $i = 1,2,\ldots,k$, where $\bigcup_{i=1}^{k}{V_{i}} = V$. The graph $\mathcal{G}$ can be partitioned by removing the edges that connect each of the subgraphs to every other subgraph. A standard partitioning cost is the \emph{multiclass normalized cut} cost \cite{yu2003msc}, defined as a generalization of \cite{shi2000nci}: 
\begin{align} \label{eq:ncutcost}
	\textrm{NCut}_{\mathbf{W}}\!\left(V_{1},\ldots,V_{k}\right) &= \sum_{i=1}^{k}{\frac{\textrm{Cut}_{\mathbf{W}}\!\left(V_{i},V\backslash V_{i}\right)}{\textrm{Vol}_{\mathbf{W}}\!\left(V_{i}\right)}} \enspace ,
\end{align}
where $\mathbf{W}$ is the weighted adjacency matrix of $\mathcal{G}$, $\textrm{Vol}_{\mathbf{W}}\!\left(V_{i}\right) = \sum_{v_{j}\in V_{i}}{d_{j}}$, and $d_{j} = \sum_{\ell}{W_{j,\ell}}$ is the degree of vertex $v_{j}$. 

By defining an $n\times k$ indicator matrix $\mathbf{X}$ so that $X_{i,j} = 1$ if $v_{i}\in V_{j}$ and $X_{i,j} = 0$ otherwise, it is straightforward to see how the multiclass normalized cut cost can be expressed in terms of Rayleigh quotients. If $\mathbf{x}_{i}$ is the $i^{\textrm{th}}$ column of $\mathbf{X}$, then $\textrm{Vol}_{\mathbf{W}}\!\left(V_{i}\right)$ can be written in terms of the degree matrix $\mathbf{D} = \textrm{diag}\!\left(\mathbf{d}\right)$ as $\mathbf{x}_{i}^{\mathbf{T}}\mathbf{D}\mathbf{x}_{i}$, and the pairwise cut cost between $V_{i}$ and $V\backslash V_{i}$ can be written as $\textrm{Cut}_{\mathbf{W}}\!\left(V_{i},V\backslash V_{i}\right) = \mathbf{x}_{i}^{\mathbf{T}}\mathbf{W}\left(\mathbf{1}-\mathbf{x}_{i}\right) = \mathbf{x}_{i}^{\mathbf{T}}\mathbf{d} - \mathbf{x}_{i}^{\mathbf{T}}\mathbf{W}\mathbf{x}_{i} = \mathbf{x}_{i}^{\mathbf{T}}\mathbf{D}\mathbf{x}_{i} - \mathbf{x}_{i}^{\mathbf{T}}\mathbf{W}\mathbf{x}_{i} = \mathbf{x}_{i}^{\mathbf{T}}\!\left(\mathbf{D}-\mathbf{W}\right) \mathbf{x}_{i}$. This allows us to express (\ref{eq:ncutcost}) as:
\begin{align} 
	\textrm{NCut}_{\mathbf{W}}\!\left(V_{1},\ldots,V_{k}\right) &= 
		\sum_{i=1}^{k}{\frac{\mathbf{x}_{i}^{\mathbf{T}}\!\left(\mathbf{D}-\mathbf{W}\right) \mathbf{x}_{i}}
			{\mathbf{x}_{i}^{\mathbf{T}}\mathbf{D}\mathbf{x}_{i}}} 
		= \textrm{tr}\!\left(\mathbf{X^{T}LX}\!\left(\mathbf{X^{T}DX}\right)^{-1}\right) \enspace , \label{eq:ncutEquiv}
\end{align}
where $\mathbf{L} = \mathbf{D}-\mathbf{W}$ is the Laplacian matrix of $\mathcal{G}$. Minimizing (\ref{eq:ncutEquiv}) is NP-hard; however, a fast approximate minimum can be found by relaxing the binary constraints on the entries of $\mathbf{X}$, minimizing the relaxed version of (\ref{eq:ncutEquiv}) by computing the generalized eigenvectors corresponding to the smallest $k-1$ nontrivial generalized eigenvalues of $\mathbf{L\hat{x}} = \lambda\mathbf{D\hat{x}}$, and discretizing the result by $k$-means clustering.

\section{Multiple-View Spectral Clustering}
\label{sec:multiView}

Now consider a collection of $m$ undirected weighted graphs $\mathcal{G}^{\left(\ell\right)} = \left(V,\mathcal{E}^{\left(\ell\right)}\right)$, $\ell = 1,\ldots, m$ that share a common set of vertices but different edges, and define $\mathbf{W}^{\left(\ell\right)}$ to be the weighted adjacency matrix for $\mathcal{G}^{\left(\ell\right)}$. Finding a common partitioning of all of the graphs, as in the single-view case, requires defining and optimizing a partitioning cost. A natural partitioning cost that generalizes (\ref{eq:ncutcost}) can be formulated in a manner that is equivalent to the spectral clustering technique of Zhou and Burges \cite{zhou2007sct}:
\begin{align} 
	\textrm{NCut}_{\mathbf{\tilde{W}}}\!\left(V_{1},\ldots,V_{k}\right) &= 
		\sum_{i=1}^{k}{\frac{\textrm{Cut}_{\mathbf{\tilde{W}}}\!\left(V_{i},V\backslash V_{i}\right)}{\textrm{Vol}_{\mathbf{\tilde{W}}}\!\left(V_{i}\right)}} 
		= \textrm{tr}\!\left(\mathbf{X^{T}\tilde{L}X}\!\left(\mathbf{X^{T}\tilde{D}X}\right)^{-1}\right) \enspace , \label{eq:mncut}
\end{align}
where $\mathbf{\tilde{W}} = \sum_{\ell=1}^{m}{\alpha_{\ell}\mathbf{W}^{\left(\ell\right)}}$, the $\alpha_{\ell}$'s are nonnegative weights that sum to one, and $\mathbf{\tilde{D}}$ and $\mathbf{\tilde{L}}$ are the degree and Laplacian matrices, respectively, of the graph having weighted adjacency matrix $\mathbf{\tilde{W}}$. As described in \cite{zhou2007sct}, this partitioning cost arises when modeling a mixture of random walkers on undirected graphs. Its minimum yields a good cut on average across the collection of graphs, even though it may not be the best cut for the underlying individual graphs in the collection. 

In practice, minimizing (\ref{eq:mncut}) is NP-hard, but as with (\ref{eq:ncutEquiv}), a relaxed version of (\ref{eq:mncut}) can be solved by identifying the generalized eigenvectors corresponding to the smallest $k-1$ nontrivial generalized eigenvalues of $\mathbf{\tilde{L}\hat{x}} = \lambda\mathbf{\tilde{D}\hat{x}}$, and the result can subsequently be discretized by $k$-means clustering.

\vspace{10pt}

\noindent
{\bf Choice of Weights:} $\quad$ The choice of $\alpha_{1}, \ldots, \alpha_{m}$ is a hyperparameter that must be determined by the user. If uniform weights are chosen ($\alpha_{j} = m^{-1}$, $j = 1, \ldots, m$), we refer to the multiple-view spectral clustering algorithm as MVSC. Another option is to include weight selection in the optimization itself, for example, via the line search technique described in Huang et al. \cite{huang2012aas}. Huang et al. refer to this approach as Affinity Aggregation for Spectral Clustering (AASC). 

We propose a third option, motivated from the idea that weights should be chosen for each individual graph in proportion to the quality of its embedding. Zhang and Jordan \cite{zhang2008msc} show that the minimum normalized cut cost for $\mathcal{G}^{\left(\ell\right)}$ is equal to the sum of the smallest $k-1$ nontrivial generalized eigenvalues satisfying $\mathbf{L}^{\left(\ell\right)}\mathbf{x} = \lambda\mathbf{D}^{\left(\ell\right)}\mathbf{x}$. If we call these eigenvalues $\lambda^{\left(\ell\right)}_{1}$, $\lambda^{\left(\ell\right)}_{2}$, $\ldots$, $\lambda^{\left(\ell\right)}_{k-1}$, then the choice of weights:
\begin{align}
	\alpha_{j} &= \frac{\left[\sum_{i=1}^{k-1}{\lambda^{\left(j\right)}_{i}}\right]^{-1}}
		{\sum_{\ell=1}^{m}{\left[\sum_{i=1}^{k-1}{\lambda^{\left(\ell\right)}_{i}}\right]^{-1}}} \enspace , \quad j = 1, \ldots, m \enspace , \label{eq:optimweights}
\end{align}
will ensure that individual graphs with high quality partitioning cost will be more heavily weighted in the computation of the group-average Laplacian than individual graphs with lower quality partitioning cost. In the sequel, we will refer to multiple-view spectral clustering with weights given by (\ref{eq:optimweights}) as MVSCW.

\vspace{10pt}

\noindent
{\bf Relationship to Joint Diagonalization of Laplacians:} $\quad$ The recent group-wise functional community detection technique proposed by Dodero et al. \cite{dodero2014gwf} identifies an orthogonal matrix $\mathbf{Q}$ that approximately jointly diagonalizes all of the normalized graph Laplacian matrices by minimizing: 
\begin{align} \label{eq:dodero}
	\varepsilon\!\left(\mathbf{Q}\right) &= \sum_{\ell=1}^{m}{\textit{off}\!\left(\mathbf{Q^{T}}{\mathbf{D}^{\left(\ell\right)}}^{-1/2}\mathbf{L}^{\left(\ell\right)}{\mathbf{D}^{\left(\ell\right)}}^{-1/2}\mathbf{Q}\right)} \enspace ,
\end{align}
subject to the orthogonality constraint $\mathbf{Q^{T}Q} = \mathbf{I}$, where $\textit{off}\!\left(\mathbf{A}\right) = \sum_{i\neq j}{\left|a_{i,j}\right|^{2}}$. 

While this approach does enable the use of a simultaneous diagonalization algorithm that has a theoretical guarantee of convergence \cite{cardoso1996jas}, the cost function (\ref{eq:dodero}) does not directly quantify the cost of partitioning the graph vertices. To see why this is true, consider the partitioned matrix $\mathbf{Q} = \left[\mathbf{Q}_{1}\,\mathbf{Q}_{2}\right]$, where $\mathbf{Q}_{1}$ is $n\times k$ and $\mathbf{Q}_{2}$ is $n\times \left(n-k\right)$. Suppose the rows of $\mathbf{Q}_{1}$ form an embedding of the graph vertices that is rounded to form the cluster indicator matrix $\mathbf{X}$. Then, in order for (\ref{eq:dodero}) to be thought of as approximately modeling some graph partitioning cost, $\varepsilon\!\left(\mathbf{Q}\right)$ should be invariant with respect to any transformation of the form $\mathbf{Q}_{2} \leftarrow \mathbf{Q}_{2}\mathbf{U}$, where $\mathbf{U}$ is any orthogonal $\left(n-k\right)\times\left(n-k\right)$ matrix. However, (\ref{eq:dodero}) does not exhibit this type of invariance.

%%%%%%%%%%%%%%%%%%%%%%%%%%%%%%%%%%%%%%%%%%%%%%%%%%%%%%%%%%%%%%%%
%%%%%%%%%%%%%%%%%%%%%%%%%%%%%%%%%%%%%%%%%%%%%%%%%%%%%%%%%%%%%%%%
\section{Experiments}
\label{sec:experiments}

In this section, we compare three different multiple-view spectral clustering algorithms for group-wise community detection (MVSC, MVSCW, AASC) with Joint Diagonalization of Laplacians (JDL). We use each of the four algorithms to compute embeddings, and then we generate labellings via $k$-means clustering. Clustering is performed 100 times with different random seeds; the mode of each vertex is selected as the label for that vertex. Since labellings are ambiguous up to permutation, every comparison between two or more labellings includes a step of identifying the permutation that maximizes the relative overlap (Dice coefficient) between labellings. 

\vspace{10pt}

\noindent
{\bf Data:} $\quad$ To compare algorithms, we utilize rsfMRI data comprising one run from each of 291 female participants from the S500 release of the Human Connectome Project (HCP) \cite{hcpOnline}. All images were collected on a 3T Siemens Skyra scanner with a 32-channel receive head coil. T2* weighted functional images were acquired using a gradient-echo EPI sequence with TE $= 33.1$ ms, TR $ = 0.72$ s, flip angle $= 52^{\circ}$, slice thickness $= 2$ mm, field of view $= 208\times 180$ mm, matrix size $= 104\times 90$, voxel size $= 2\times 2\times 2$ mm$^{3}$. Scan duration was approximately 15 minutes, yielding $\sim 1200$ volumes. HCP provides preprocessed rsfMRI data \cite{glasser2013mpp}; we select the data that has been subsequently post-processed according to the FIX protocol \cite{salimi2014adf} for ICA-based denoising.

Each FIX-preprocessed rsfMRI is parcellated according to the Automatic Anatomical Labeling (AAL) atlas \cite{tzourio2002aal}, which divides the brain into 90 cortical/subcortical regions and 26 cerebellar/vermis regions. Time series from voxels within each region are averaged to form a representative time series for each region. The weighted adjacency matrices $\mathbf{W}^{\left(\ell\right)}$ are constructed from Fisher $z$-transformed Pearson correlation coefficients between the representative time series for each region. Negative weights are zeroed, ensuring all graph Laplacian matrices remain positive semi-definite.

\vspace{10pt}

\noindent
{\bf Number of Clusters:} $\quad$ The true underlying number of functional clusters in the brain is unknown. For these experiments, 
we choose the number of clusters based on the presence of an eigenvalue gap in the group-averaged Laplacian when using the full set of 291 female brains. As shown in Figure 1, when the MVSC and JDL algorithms are used, there is a large gap between the $4^\textrm{th}$ and $5^\textrm{th}$ eigenvalues and a smaller gap between the $7^{\textrm{th}}$ and $8^{\textrm{th}}$ eigenvalues. For this reason, we compute two versions of group-averaged Laplacians for use in the MVSCW algorithm: one with the choice $k=5$, and the other with $k=8$. (Note that the weights in MVSCW depend on the number of desired clusters.) For both of these cases, MVSCW exhibits similar gaps between the $4^{\textrm{th}}$ and $5^{\textrm{th}}$ and between the $7^{\textrm{th}}$ and $8^{\textrm{th}}$ eigenvalues. When the AASC algorithms are used, there is a gap between the $7^{\textrm{th}}$ and $8^{\textrm{th}}$ eigenvalues, but no discernible gap between the $4^{\textrm{th}}$ and $5^{\textrm{th}}$ eigenvalues. The same eigenvalue gaps are apparent when analyzing a similar set of 203 male brains. For these reasons, we perform subsequent experiments for both $k=5$ and $k=8$ clusters. The choice of eight clusters is consistent with number of clusters found by the co-training algorithm of Chen et al. \cite{chen2013igw}.
 
\begin{figure}[p]
	\centering
		\includegraphics[scale=0.68]{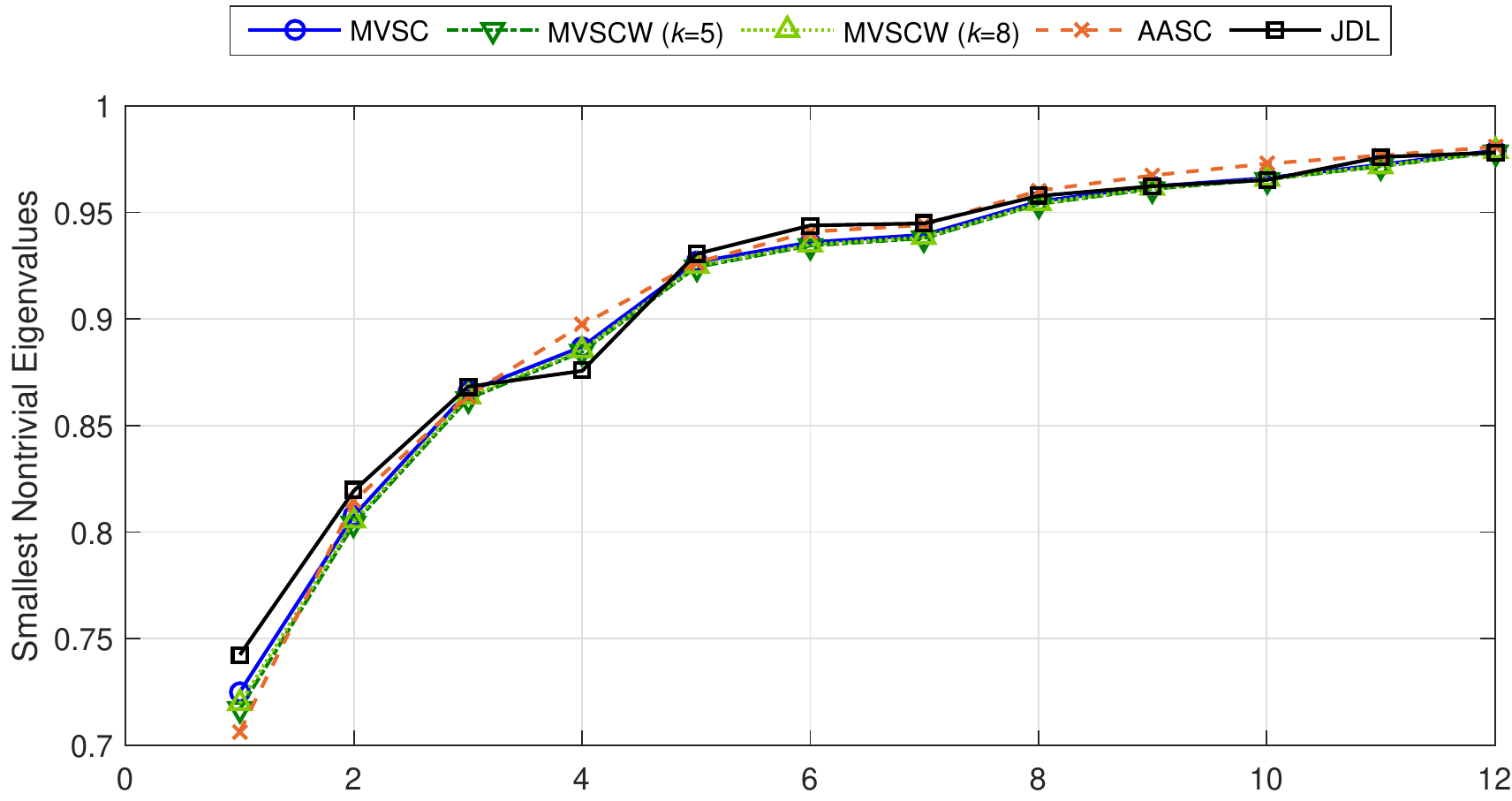}
	\label{fig:eigenvalueFigure}
	\caption{Smallest nontrivial generalized eigenvalues computed on 291 female brains using MVSC, MVSCW, AASC, and JDL algorithms. A large gap is apparent between the $4^{\textrm{th}}$ and $5^{\textrm{th}}$ eigenvalues for MVSC, MVSCW, and JDL, and a smaller gap is apparent between the $7^{\textrm{th}}$ and $8^{\textrm{th}}$ eigenvalues for all algorithms.}
\end{figure}

\begin{figure}[p]
\begin{center}
	\includegraphics[scale=0.72]{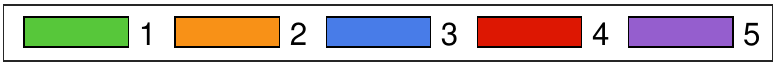}\\
	\vspace{10pt}
	\subfigure[MVSC]{\includegraphics[scale=0.64]{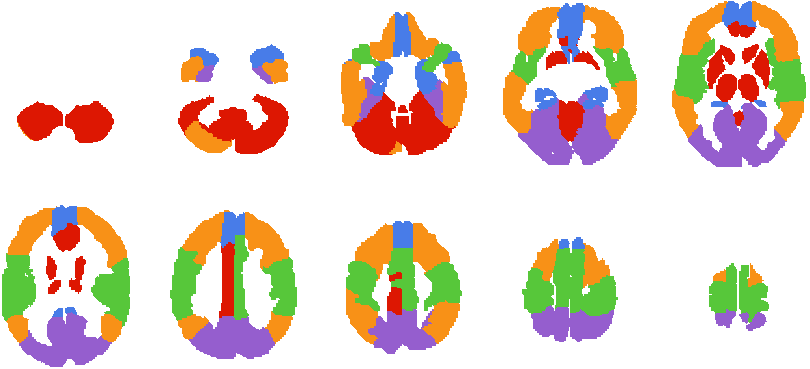}}
	\hfill
	\subfigure[MVSCW]{\includegraphics[scale=0.64]{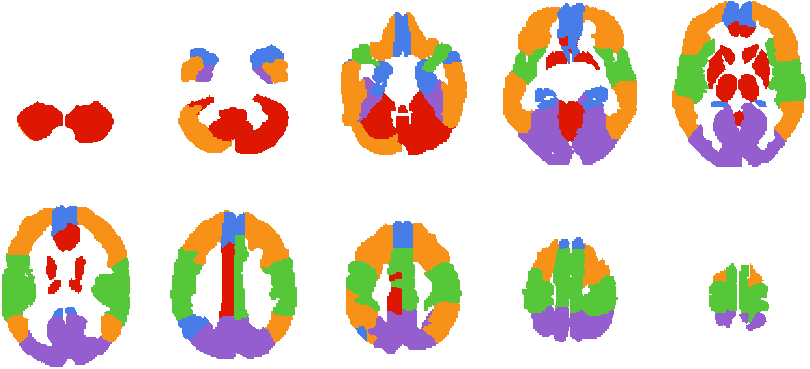}}\\
	\vspace{6pt}
	\subfigure[AASC]{\includegraphics[scale=0.64]{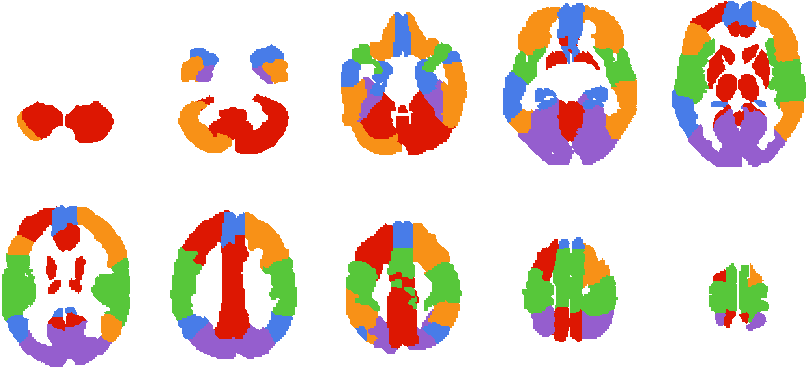}}
	\hfill
	\subfigure[JDL]{\includegraphics[scale=0.64]{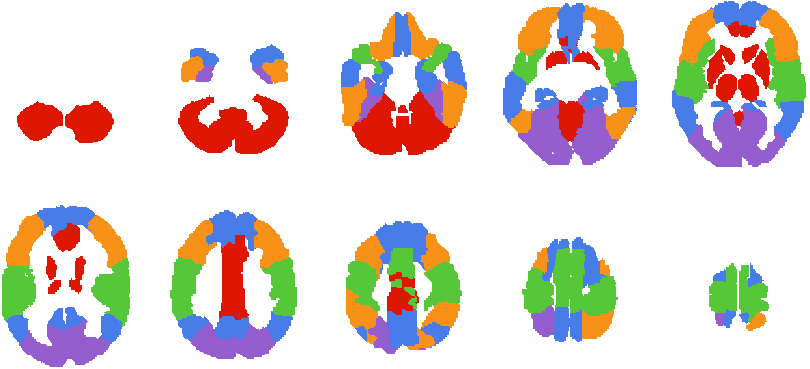}}
\end{center}
\caption{Group-wise clusters ($k=5$). Clusters roughly correspond to: 1) Sensorimotor network, 2) Default mode network (MVSC/MVSCW), 3) Orbitofrontal cortex network (MVSC/MVSCW), 4) Basal ganglia + cerebellum, and 5) Visual network. For AASC/JDL, clusters 2 and 3 roughly correspond to the union of the orbitofrontal cortex and default mode networks.}
\label{fig:clustersK5}
\end{figure}

\begin{figure}[t]
\begin{center}
	\includegraphics[scale=0.72]{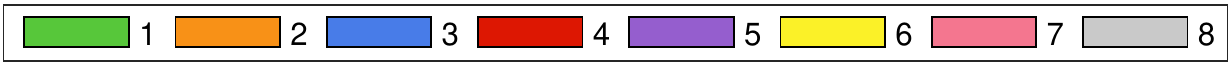}\\
	\vspace{10pt}
	\subfigure[MVSC]{\includegraphics[scale=0.64]{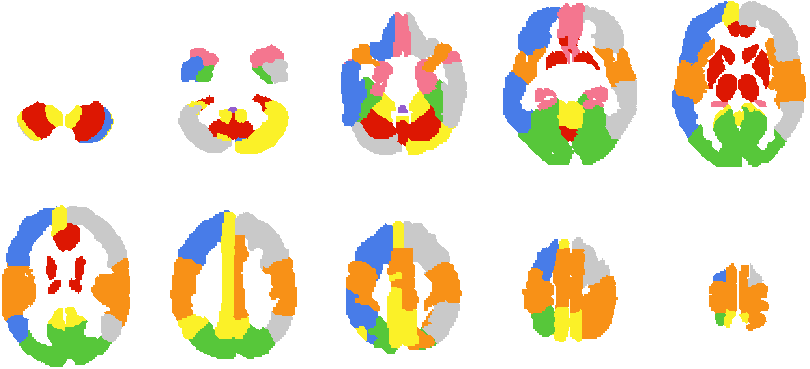}}
	\hfill
	\subfigure[MVSCW]{\includegraphics[scale=0.64]{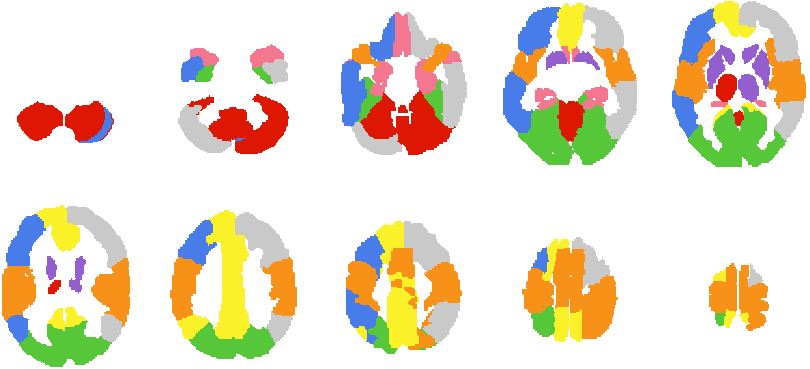}}\\
	\vspace{6pt}
	\subfigure[AASC]{\includegraphics[scale=0.64]{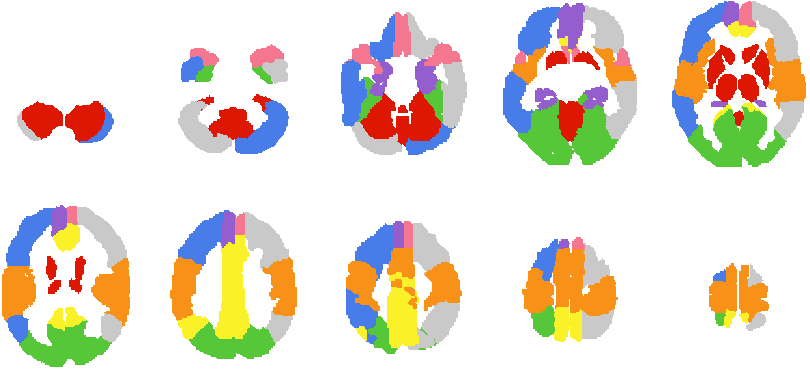}}
	\hfill
	\subfigure[JDL]{\includegraphics[scale=0.64]{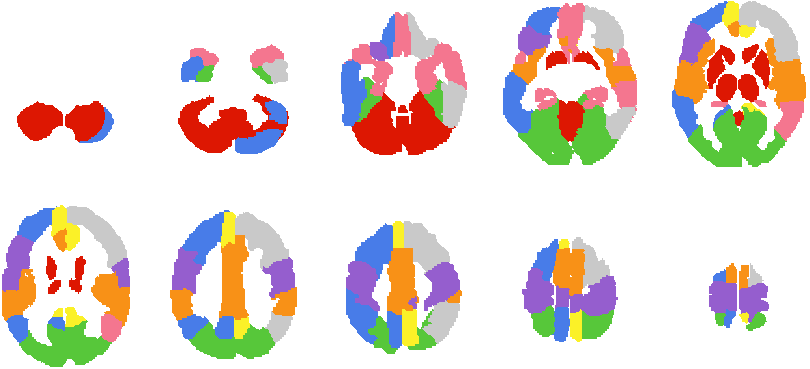}}
\end{center}
\caption{Group-wise clusters ($k=8$). Some of the identified clusters roughly correspond to: 1) Visual network, 2) Sensorimotor network, 3) Left frontoparietal network, 4) Basal ganglia + cerebellum, 7) Orbitofrontal cortex network, and 8) Right frontoparietal network.}
\label{fig:clustersK8}
\end{figure}

\vspace{10pt}

\noindent
{\bf Clustering Results:} $\quad$ Figures \ref{fig:clustersK5}--\ref{fig:clustersK8} illustrate the group-wise functional communities that result from applying MVSC/MVSCW/AASC/JDL + $k$-means clustering to the 291 female brains for the cases of $k=5$ and $k=8$ clusters. By inspecting the regions of the brain that are clustered into similar communities, we see that for the $k=5$ case, communities roughly correspond to the sensorimotor network, the default mode network, the visual network, and the combined basal ganglia and cerebellar structures. However, for AASC and JDL, the default mode and orbitofrontal cortex networks are not as nicely separated as they are by MVSC and MVSCW. For the $k=8$ case, communities emerge that roughly correspond to the sensorimotor, visual, and left and right frontoparietal networks, and the combined basal ganglia and cerebellar structures. The default mode network does not appear to emerge as a unique community for any of the algorithms for the $k=8$ case, and some of the communities (5,6) do not appear to correspond to recognizable networks.

\vspace{10pt}

\begin{figure}[t]
\centering
\includegraphics[scale=0.72]{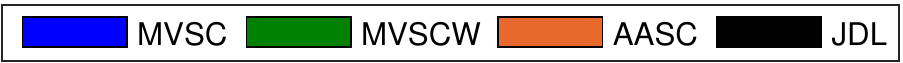}\\
\vspace{6pt}
\includegraphics[scale=0.68]{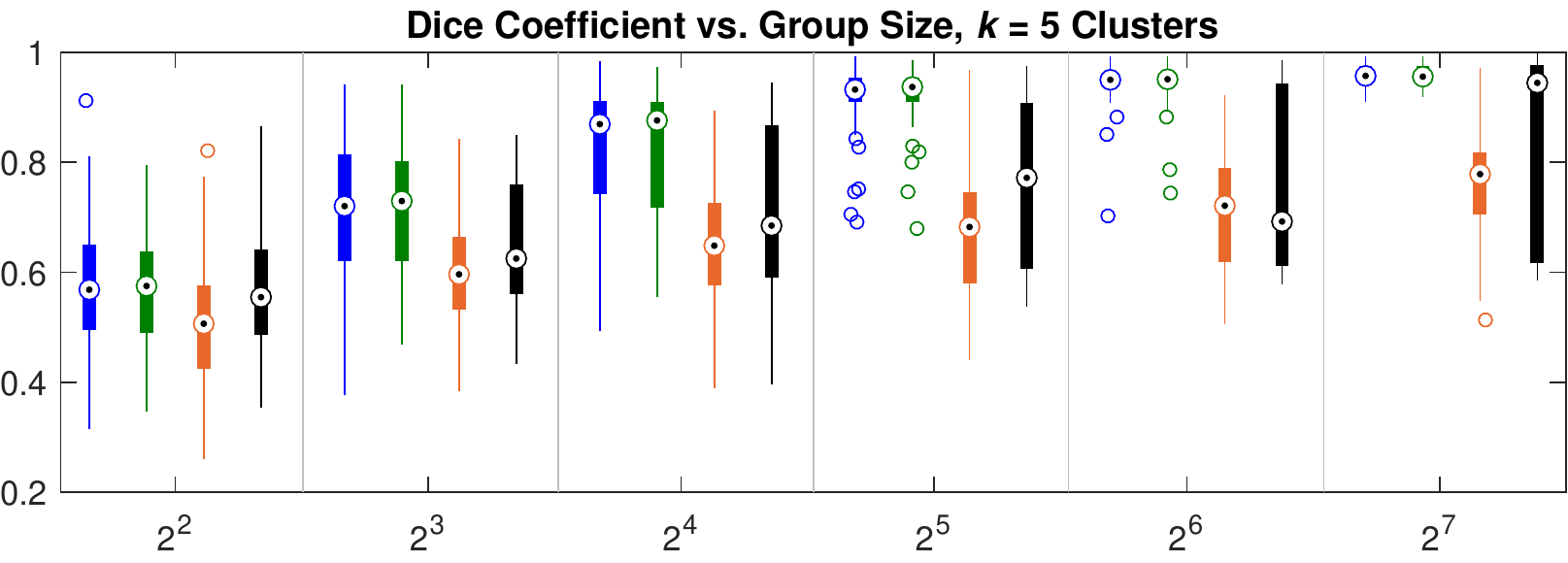}\\
\includegraphics[scale=0.68]{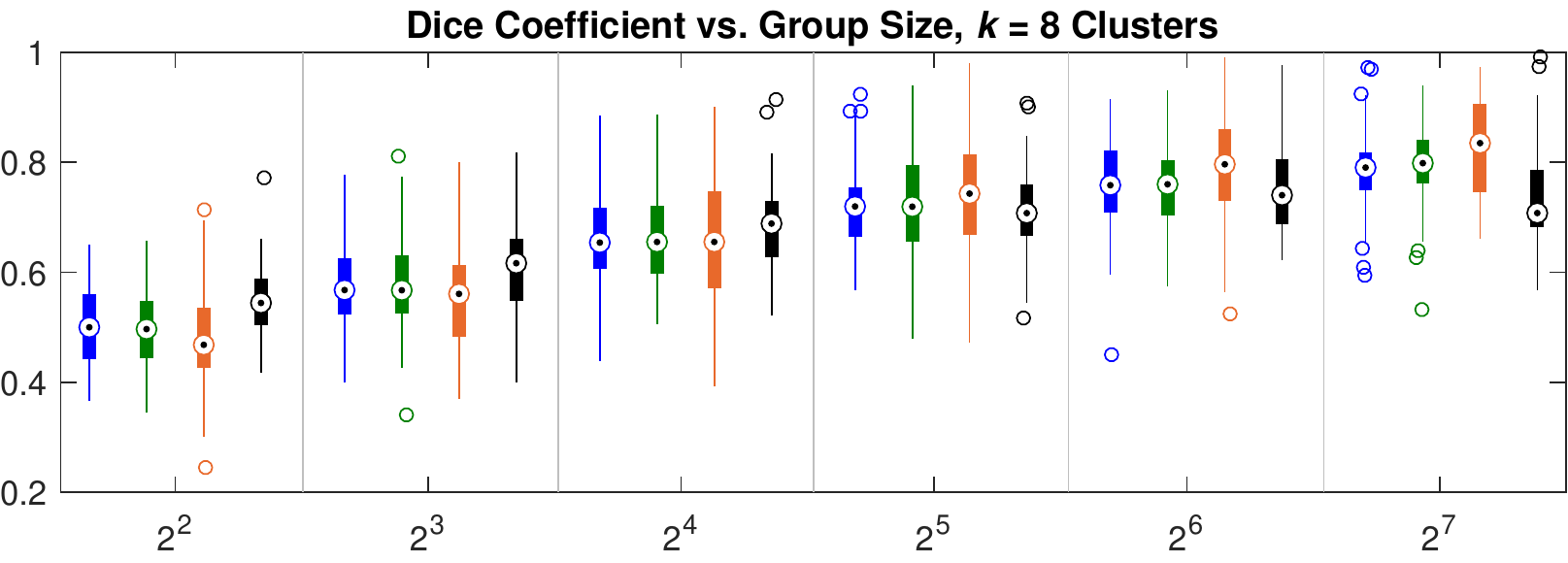}
\caption{Box plots of Dice coefficients (relative overlap in labellings) between pairs of labellings generated from MVSC/MVSCW/AASC/JDL + $k$-means clustering using groups of brains (group sizes $4$, $8$, $\ldots$, $128$) randomly drawn from the 291 female brains. }
\label{fig:diceCoeffs}
\end{figure}

\noindent
{\bf Consistency:} $\quad$ Even though it is impossible to intuit the ``correct'' clustering of this data into functional communities, it is possible to assess how consistently each algorithm yields similar communities. To compare the consistency of the MSVC, MVSCW, AASC, and JDL algorithms (with $k$-means clustering applied to each embedding), we carry out the following experiment: For group sizes of $\gamma = 4$, $8$, $16$, $32$, $64$, and $128$, we randomly select two non-overlapping subsets, each containing $\gamma$ of the 291 brains. For each subset in a pair, we compute functional communities using MSVC, MVSCW, AASC, and JDL embeddings followed by $k$-means clustering with 100 random seeds. Then, for each algorithm, we compute the Dice coefficient (relative overlap) between the labellings generated from the two subsets in the pair (after identifying the permutation of one of the labellings that gives maximal overlap
with the other labeling). This entire process is performed 100 times, both for $k=5$ and $k=8$ clusters. Box plots of the resulting Dice coefficients are shown in Figure \ref{fig:diceCoeffs}.

From the results of these experiments, we can see that for $k=5$ clusters, MVSC and MVSCW yield more consistent labellings than AASC and JDL across all group sizes. For $k=8$ clusters, JDL is more consistent than MVSC, MVSCW, and AASC for group sizes up to $16$, and AASC is the least consistent for these small group sizes. For group sizes greater than $16$, however, JDL becomes the least consistent and AASC the most consistent. For both $k=5$ and $k=8$, MVSC and MVSCW seem to exhibit similar consistency.  

\vspace{10pt}

\begin{figure}[t]
	\centering
		\includegraphics[scale=0.72]{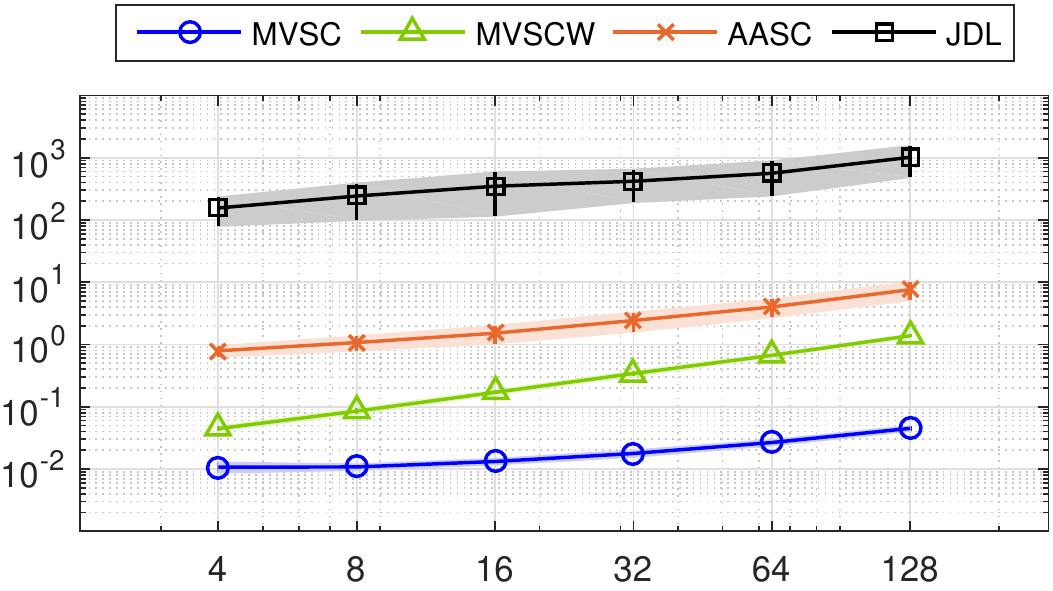}
	\label{fig:timing}
	\caption{Computation time (in seconds) required to compute MVSC, MVSCW, AASC, and JDL embeddings with $k=8$ for various group sizes. Shaded regions indicate $\pm$ one standard deviation of the mean timing result. }
\end{figure}

\noindent
{\bf Timing:} $\quad$ We would expect that of the four algorithms, MVSC, MVSCW, and AASC should be faster than JDL due to the fact that they only need to compute $k$ eigenvalues, whereas JDL must compute all $n$ eigenvalues. MVSC should be faster than MVSCW due to the weight computation in MVSCW, and it should be faster than AASC, since AASC successively solves the same subproblem that MVSC solves once. We implemented the MVSC and MVSCW algorithms in MATLAB R2015b, and we used the MATLAB implementations of AASC and JDL that are provided by their respective authors. When running the cluster consistency experiment on our quad-core desktop, we captured timing results for each algorithm, and we show the results in Figure \ref{fig:timing}. Timing results reflect the amount of time required to compute the group-wise embedding that is subsequently input into the $k$-means clustering algorithm, given the set of individual graph weighted adjacency matrices (we chose $k=8$ clusters for this experiment). As seen in the figure, MVSC is four orders of magnitude faster than JDL, two orders of magnitude faster than AASC, and one order of magnitude faster than MVSCW across all group sizes, and each algorithm appears to grow linearly in complexity with respect to the number $m$ of component graphs in the group.   

%%%%%%%%%%%%%%%%%%%%%%%%%%%%%%%%%%%%%%%%%%%%%%%%%%%%%%%%%%%%%%%%
%%%%%%%%%%%%%%%%%%%%%%%%%%%%%%%%%%%%%%%%%%%%%%%%%%%%%%%%%%%%%%%%
\section{Conclusion}
\label{sec:conc}

When modeled as a graph partitioning problem, group-wise functional community detection from brain fMRI can be carried out by performing multiple-view spectral clustering algorithms. Experiments on a set of 291 female brains show that MVSC/MVSCW/AASC yield functional communities that roughly correspond to a variety of known functional networks in the brain. In addition, when compared to the previously proposed joint diagonalization of Laplacians (JDL) technique, multiple-view spectral clustering can yield more consistent results with much faster computation times. 

%%%%%%%%%%%%%%%%%%%%%%%%%%%%%%%%%%%%%%%%%%%%%%%%%%%%%%%%%%%%%%%%
%%%%%%%%%%%%%%%%%%%%%%%%%%%%%%%%%%%%%%%%%%%%%%%%%%%%%%%%%%%%%%%%
\section*{Appendix}

Prototype implementations of the MVSC and MVSCW algorithms, as well as wrappers around the original authors' implementations of AASC and JDL, are available for download at MATLAB Central \cite{matlabCentral} under File ID \#58753.

\newpage
\bibliographystyle{splncs03}
\bibliography{references}

\end{document}